\documentclass[letterpaper, 10 pt, journal, twoside]{IEEEtran}
\usepackage{amsmath}
\usepackage{booktabs}
\usepackage{graphicx}
\usepackage{amssymb}
\usepackage{xcolor}
\usepackage{float}
\usepackage{placeins}
\usepackage{threeparttable}
\usepackage{tabularx} 
\usepackage[caption=false,font=footnotesize]{subfig}
\bstctlcite{IEEEexample:BSTcontrol}
\title{RobotDancing: Residual-Action Reinforcement Learning Enables Robust Long-Horizon Humanoid Motion Tracking}

\author{Zhenguo Sun$^{1,2,3}$, Yibo Peng$^{2}$, Yuan Meng$^{1,2}$, Xukun Li$^{1,2,3}$, Yu Sun$^{2}$,
Haojun Jiang$^{4}$, Bo-Sheng Huang$^{2,5}$, Zhenshan Bing$^{6}$,~\IEEEmembership{Member,~IEEE}, Xinlong Wang$^{2}$,
and Alois Knoll$^{1}$,~\IEEEmembership{Fellow,~IEEE}%
\thanks{Manuscript received April 5, 2026; revised July 3, 2026; accepted August 1, 2026.
This paper was recommended for publication by Editor Olivier Stasse upon evaluation of the Associate Editor and Reviewers' comments.
\textit{Zhenguo Sun and Yibo Peng contributed equally.}
\textit{Corresponding authors: Zhenshan Bing and Xinlong Wang.}}%
\thanks{$^{1}$Zhenguo Sun, Yuan Meng, and Alois Knoll are with the School of Computation, Information and Technology, Technical University of Munich, 85748 Munich, Germany {\tt\footnotesize hitsunzhenguo@gmail.com}, {\tt\footnotesize y.meng@tum.de}, {\tt\footnotesize k@tum.de}.

$^{2}$Yibo Peng, Yu Sun, and Xinlong Wang are with the Beijing Academy of Artificial Intelligence, 100084 Beijing, China {\tt\footnotesize ybpeng@baai.ac.cn}, {\tt\footnotesize sunyu@baai.ac.cn}, {\tt\footnotesize xlwang\_1@baai.ac.cn}.

$^{3}$Xukun Li is with XYZ Embodied AI, 100080 Beijing, China {\tt\footnotesize xukunli0221@gamil.com}.

$^{4}$Haojun Jiang is with the Department of Automation, Tsinghua University, 100084 Beijing, China {\tt\footnotesize jianghaojun2015@gmail.com}.

$^{5}$Bo-Sheng Huang is with the Department of Computer Science, Tsinghua University, 100084 Beijing, China {\tt\footnotesize bosheng791@gmail.com}.

$^{6}$Zhenshan Bing is with the State Key Laboratory for Novel Software Technology, Nanjing University, 215163 Suzhou, China {\tt\footnotesize bing@nju.edu.cn}.}%

\thanks{Digital Object Identifier (DOI): see top of this page.}
}

\begin{document}

\maketitle

\begin{abstract}
Long-horizon, high-dynamic motion tracking on humanoids remains brittle: retargeted reference motions are typically kinematically plausible but dynamically inconsistent with the robot, so small tracking errors accumulate and eventually destabilize control.
We present RobotDancing, a practical single-stage reinforcement learning recipe that tracks retargeted motions by predicting reference-conditioned residual joint targets.
By parameterizing actions as residual corrections on top of the reference, the policy can focus on compensating reference--robot dynamics mismatch (e.g., actuation limits, latency, friction, inertia) rather than re-synthesizing the motion.
We train one policy per reference sequence while reusing a common training and deployment recipe across motions and platforms.
We evaluate RobotDancing on eight LAFAN1 dance motions on Unitree G1 and conduct cross-platform experiments on H1 and H1-2.
The resulting G1 policies execute long-horizon, high-energy behaviors on hardware without test-time modifications, and qualitative H1/H1-2 clips further illustrate cross-platform feasibility.
\end{abstract}


\section{INTRODUCTION}

\IEEEPARstart{H}{umanoid} robots are increasingly expected to execute long-horizon, highly dynamic behaviors such as dance, where small tracking errors compound rapidly and destabilize control.
A principal source of long-horizon drift is reference--robot dynamics mismatch: the mismatch between reference trajectories and the robot's true physics (actuation limits, friction, inertia, latency).
While the animation community has shown impressive progress in whole-body control of physics-based characters~\cite{deepmimic2018, luo2023universal, tessler2024maskedmimic}, precise and robust motion tracking remains a core challenge in humanoid robotics, especially when transferring motions from simulation or captured human data to real robotic platforms.

Recent advances in physics-based humanoid control~\cite{pbhc2025,asap2025,exbody2024,omnih2o2024,fu2024humanplus,chen2025gmt,yin2025unitracker,wang2025experts} demonstrate that learning-based controllers can produce agile behaviors under physical constraints.
Nevertheless, many motion-tracking systems still rely on predicting absolute joint commands.
This works well for short or quasi-cyclic skills, but often becomes fragile on long, high-energy sequences: accumulated discrepancies between the reference and the robot's actual dynamics gradually grow, leading to drift and eventual failure.

Inspired by residual learning paradigms~\cite{ctrl2024, luo2024residual}, RobotDancing addresses this practical gap with a simple design choice: we track by correcting.
Instead of learning absolute joint targets, the policy outputs residual corrections on top of the reference, preserving its kinematic structure while focusing learning on dynamics compensation.
To improve long-horizon training efficiency, we combine distribution-aware balancing with failure-aware prioritization to cover rare poses and revisit persistently difficult segments.

RobotDancing spans training, sim-to-sim verification, and zero-shot sim-to-real deployment.
We train one policy per reference sequence; RobotDancing is therefore a reusable per-sequence recipe rather than a universal tracker.
We evaluate eight retargeted LAFAN1 dance motions primarily on Unitree G1 and conduct cross-platform experiments on H1 and H1-2.
The primary contributions are:
\begin{itemize}
    \item \textbf{A reproducible per-sequence motion-tracking recipe.}
    We present a single-stage pipeline for long-horizon, high-dynamic motion tracking.

    \item \textbf{Reference-conditioned selective residual actions for dynamics compensation.}
    The policy outputs residual corrections to the reference joint targets online, reducing error accumulation and improving stability.

    \item \textbf{Effective sampling for long-tail motions.}
    Distribution-aware balancing and failure-aware prioritization improve coverage and focus learning on difficult segments.

    \item \textbf{Evaluation across motions and platforms.}
    We validate long-horizon tracking on eight dance motions and conduct experiments on G1, H1, and H1-2.
\end{itemize}

\section{RELATED WORK}

\begin{figure*}[!t]
  \centering
  \includegraphics[width=\textwidth]{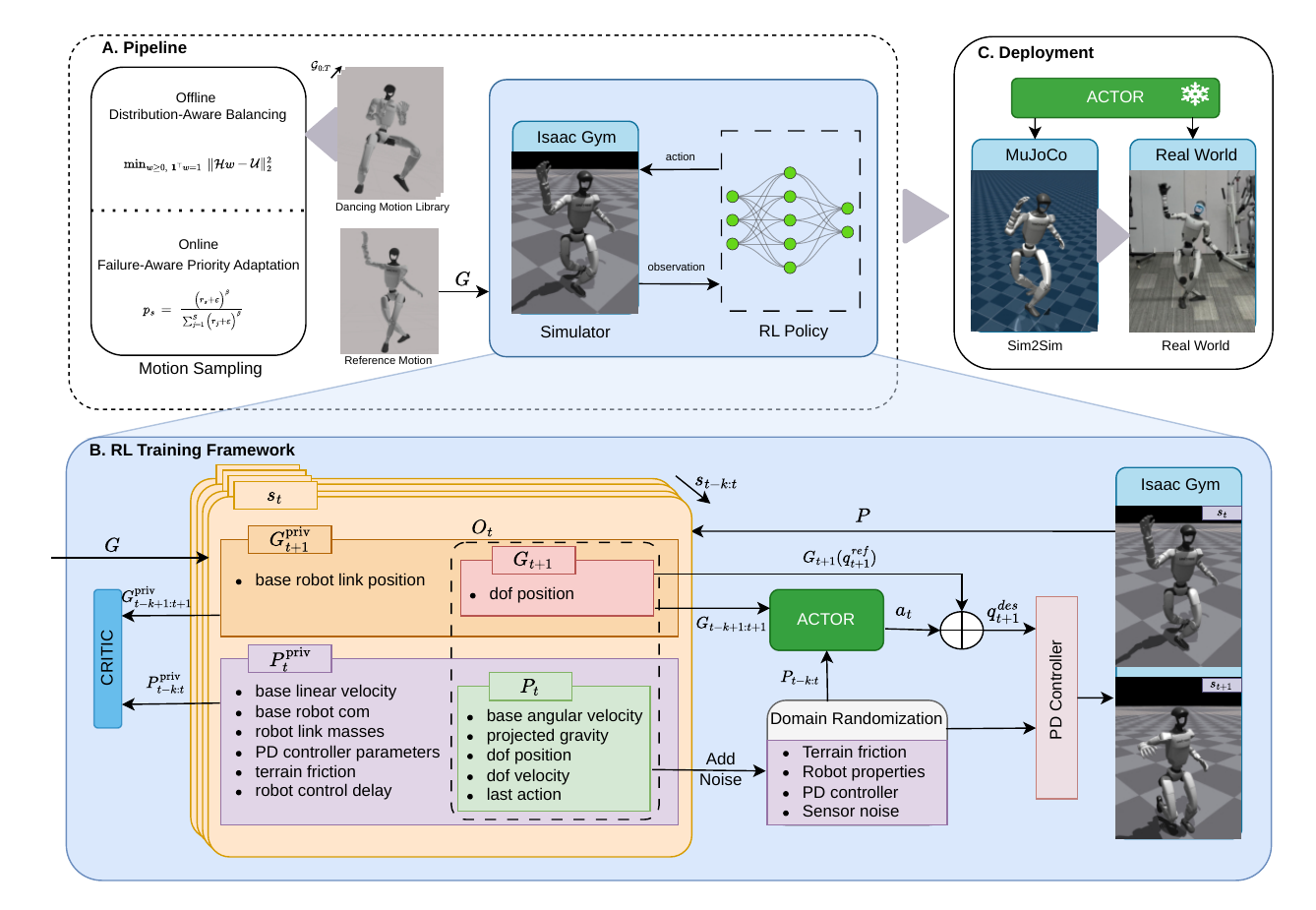} 
  \caption{\textbf{Framework of RobotDancing.} (a) The pipeline from motion data processing to RL policy learning. (b) The RL training framework for long-horizon high-dynamic motion tracking tasks through residual action learning. (c) The deployment framework enables Sim2Sim and Sim2Real transfers.}
  \label{fig:framework}
\end{figure*}

\subsection{Motion Tracking for Humanoids}
Learning-based control has achieved impressive results on task-specific humanoid skills such as locomotion, balancing, and getting up~\cite{he2025attention,zhang2025hub, huang2025host,radosavovic2024real}, yet these controllers typically rely on careful reward engineering and often yield motions that are not consistently human-like.
By contrast, motion tracking seeks to reproduce human demonstrations, providing a direct route to natural whole-body behaviors.

\textit{Single-skill tracking.}
DeepMimic~\cite{deepmimic2018} established the canonical formulation in simulation, demonstrating that reference-conditioned policies can reproduce diverse skills under physics constraints. 
ASAP~\cite{asap2025} aligns simulation and real-world physics for agile skills, while KungfuBot~\cite{pbhc2025} combines whole-body tracking with curriculum and domain randomization for highly dynamic behaviors.

\textit{General motion tracking and teleoperation.}
Recent systems pursue broader motion coverage: ExBody2~\cite{exbody2024} uses teacher--student learning for expressive tracking; OmniH2O~\cite{omnih2o2024} combines teleoperation and learning; PHC~\cite{luo2023phc} emphasizes recovery; and GMT~\cite{chen2025gmt} and UniTracker~\cite{yin2025unitracker} target general humanoid tracking.

\textit{Long-horizon tracking.}
Long-horizon tracking remains challenging because dynamics mismatch accumulates over time.
BeyondMimic~\cite{truong2025beyondmimic} addresses sequence-level execution through generation and sampling, whereas RobotDancing focuses on online residual correction of retargeted references.

\subsection{Residual Learning in Robotics}
Residual RL refines nominal controllers and has improved contact-rich manipulation and locomotion~\cite{luo2024residual,johannink2019residual,xie2018feedback}.
Closest to our scope, I-CTRL~\cite{ctrl2024} uses bounded residual actions for humanoid motion tracking, demonstrating that residualization can improve imitation feasibility.

\subsection{Discussion and Our Focus}
Retargeted human motions can be kinematically plausible yet dynamically inconsistent with the robot, making absolute-action tracking fragile over long horizons.
Across representative methods, DeepMimic-style trackers use reference-conditioned absolute actions for individual skills; ASAP and KungfuBot emphasize agile hardware skills through alignment and curriculum; BeyondMimic targets sequence-level execution; GMT and UniTracker broaden policy-level motion coverage; and I-CTRL uses bounded residual actions. RobotDancing instead trains one policy per long reference and combines reference-conditioned residual targets, selective residual authority, and distribution-/failure-aware sampling for simulation-to-hardware tracking. Together, these choices form a simple per-sequence recipe for long-horizon, high-dynamic motion tracking, validated across eight dance motions and three humanoid platforms.

\section{METHOD}
\begin{figure*}[t]
  \centering
  \newcommand{\w}{0.32\textwidth}
  \newcommand{\gap}{0.01\textwidth}
  \subfloat[\textit{H1}: high-dynamic dance steps with fast hip/knee pitch swings and torso stabilization.]{%
    \includegraphics[width=\w]{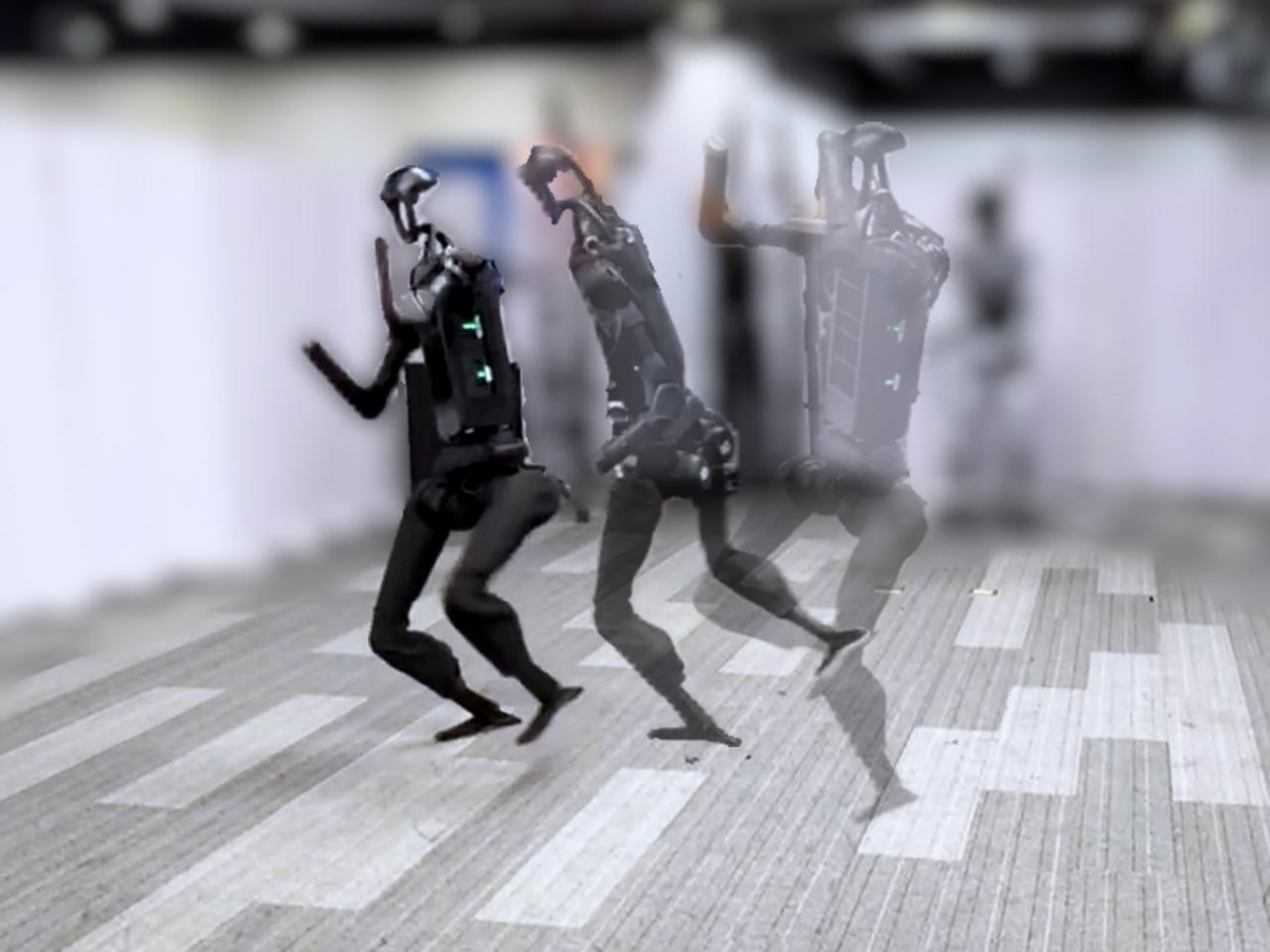}}\hspace{\gap}%
  \subfloat[\textit{G1}: rapid rhythm changes and coordinated arm--leg gestures.]{%
    \includegraphics[width=\w]{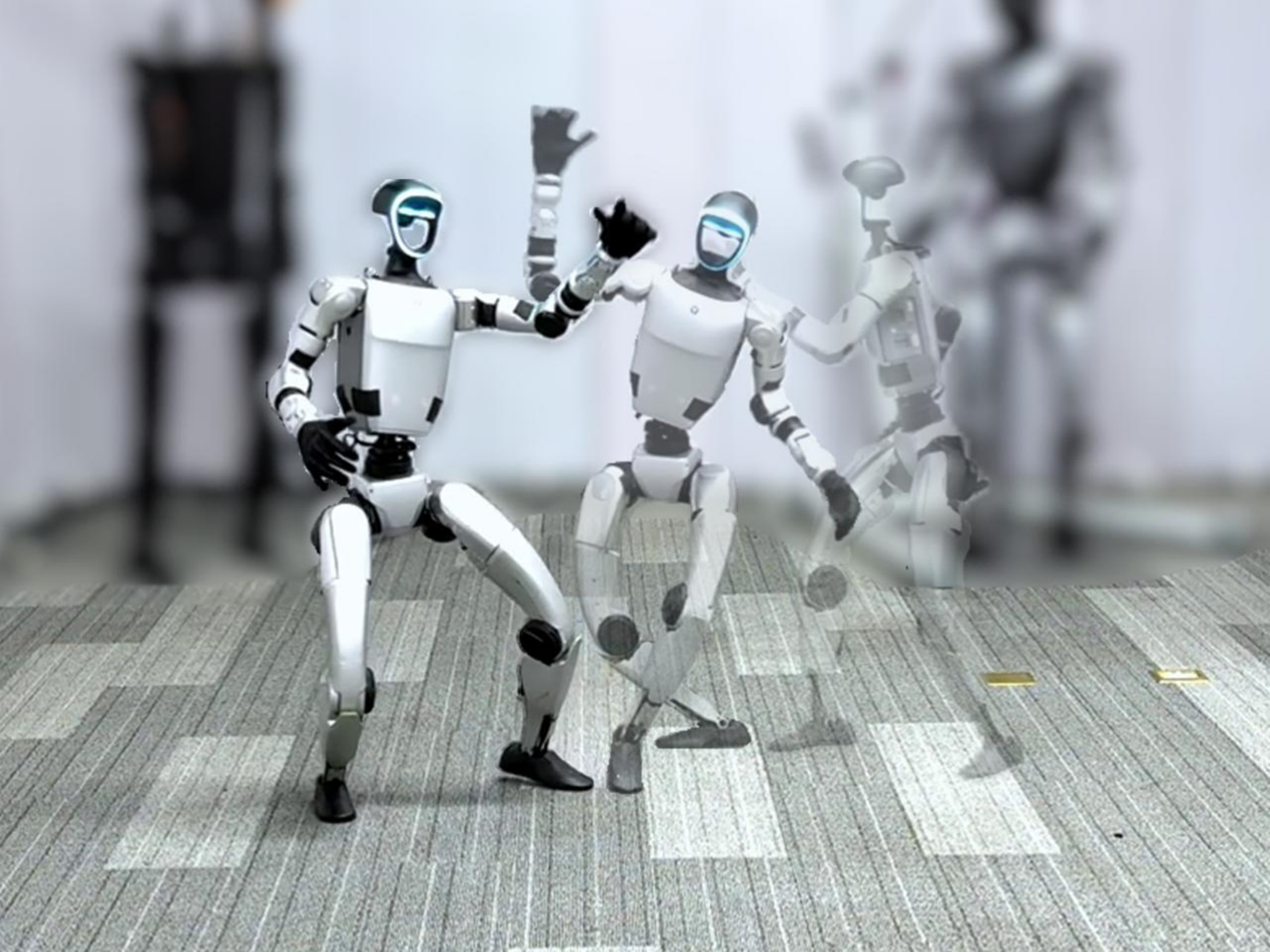}}\hspace{\gap}%
  \subfloat[\textit{H1-2}: large-amplitude lower-body motions and fast heading changes.]{%
    \includegraphics[width=\w]{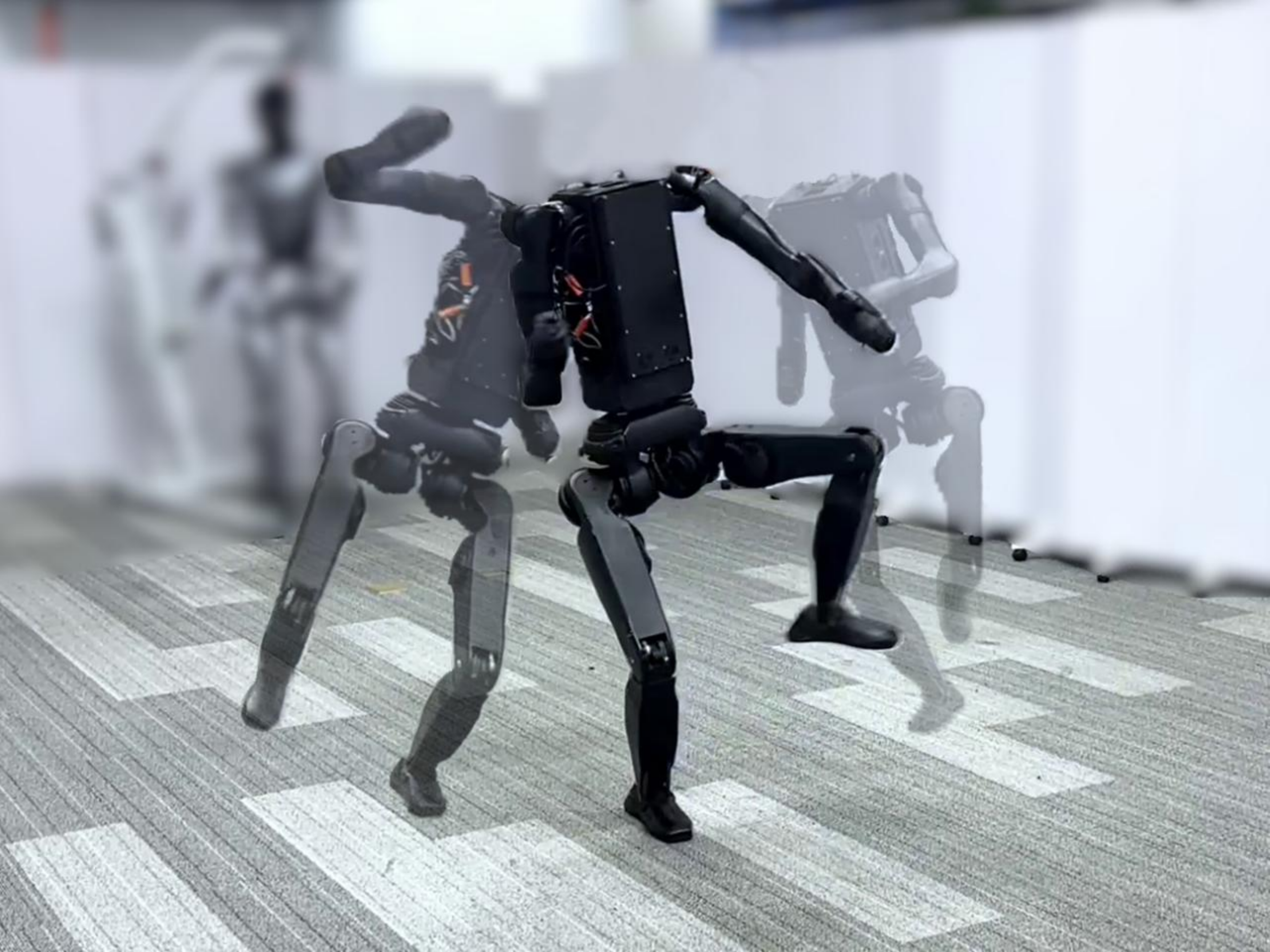}}
  \caption{\textbf{Real-World High-Dynamic Motions across Platforms.}
  Multiple platforms perform long-horizon, high-dynamic dance sequences.
  The same residual-action training recipe is applied across platforms with morphology-specific robot configurations.}
  \label{fig:real_dance_multi}
\end{figure*}

We present \textbf{RobotDancing}, a reusable per-sequence training recipe for long-horizon, high-dynamic motion tracking.
As illustrated in Fig.~\ref{fig:framework}, a goal-conditioned policy corrects reference joint targets, and a low-level PD controller executes the resulting commands.
Training combines distribution- and failure-aware resampling, selective residual actions, curriculum learning, and domain randomization.

\subsection{Problem Formulation and Notation}
\label{subsec:notation}

We formulate motion tracking as a discrete-time POMDP with state $s_t\in\mathcal{S}$, observation space $\mathcal{O}$, action space $\mathcal{A}$, reference trajectory $G_{0:T}$, and transition $p(s_{t+1}\mid s_t,a_t)$.
Proprioception $P_t$ contains onboard measurements, while $G_{t+1}$ provides the next reference joint configuration and kinematic context.
The actor observation is
\begin{equation}
  O_t=\Psi(P_t,G_{t+1})\triangleq[P_t;G_{t+1}]\in\mathcal{O},
  \label{eq:observation}
\end{equation}
where $\Psi$ denotes concatenation.
The action $a_t\in\mathbb{R}^{d_{\mathrm{act}}}$ is a residual joint-position target, with $d_{\mathrm{act}}\in\{23,21,19\}$ for G1/H1-2/H1, respectively.

\subsection{RL Training Framework}
\label{subsec:rlframework}

\textbf{Goal-Conditioned Policy.}
Using short history windows $o_t=P_{t-k:t}$ and $g_t=G_{t-k+1:t+1}$, the policy is
\begin{equation}
  a_t\sim\pi_\theta(a_t\mid o_t,g_t).
  \label{eq:policy}
\end{equation}
Following standard asymmetric actor--critic practice~\cite{deepmimic2018,pbhc2025,asap2025}, the actor receives joint states, projected gravity, base angular velocity, previous actions, and reference joints. The critic additionally receives privileged base velocity, reference link positions, and randomized physical parameters. Only actor observations are required at deployment.

\textbf{Residual Actuation.}
The policy output corrects the next reference target before joint-space PD control:
\begin{align}
  q^{\mathrm{tar}}_t &= q^{\mathrm{ref}}_{t+1}+a_t,
  \label{eq:residual}\\
  \tau_t &= \mathrm{PD}(q^{\mathrm{tar}}_t,q_t,\dot q_t).
  \label{eq:pd}
\end{align}
The reference retains the kinematic structure of the motion, while the bounded correction focuses policy capacity on reference--robot dynamics mismatch caused by actuation, contact, latency, and retargeting errors.

\textbf{Rewards.}
We separate tracking quality from physical regularization:
\begin{equation}
  r_t=r_t^{\mathrm{track}}-s_{\mathrm{pen}}(t)r_t^{\mathrm{reg}},
  \label{eq:reward_total}
\end{equation}
where $s_{\mathrm{pen}}(t)$ is curriculum controlled.
Following DeepMimic-style shaping~\cite{deepmimic2018}, tracking terms use Gaussian kernels:
\begin{equation}
  r_t^{\mathrm{track}}=
  \sum_{\chi\in\mathcal{C}}w_\chi
  \exp\!\left(-\bar e_\chi(t)/\sigma_\chi^2\right),
  \label{eq:reward_tracking}
\end{equation}
where $\mathcal{C}$ covers body, root, feet, keypoint, joint, velocity, and contact targets.
Physical regularization is summarized as
\begin{equation}
\begin{aligned}
  r_t^{\mathrm{reg}}={}&
  \lambda_\tau\|\tau_t\|_2^2+
  \lambda_{\Delta a}\|a_t-a_{t-1}\|_2^2\\
  &+\lambda_{\mathrm{lim}}\mathcal{P}_{\mathrm{limits}}
  +\lambda_{\mathrm{cnt}}\mathcal{P}_{\mathrm{contacts}}
  +\lambda_{\mathrm{term}}\mathbb{1}_{\mathrm{term}} .
\end{aligned}
  \label{eq:reward_reg}
\end{equation}
As in prior work~\cite{pbhc2025,xie2025humanoid}, we retain one reward component per value head, compute GAE independently, and sum the head-wise advantages for the PPO~\cite{schulman2017proximal} objective.

\begin{table*}[t]
\centering
\small
\setlength{\tabcolsep}{4pt}
\caption{Representative Reward Scales.}
\label{tab:method_reward_scales}
{
\begin{tabular}{lll}
\toprule
Group & Term & Scale / parameter\\
\midrule
Tracking & Body / relative body position & $2.0$ / $0.5$\\
Tracking & VR keypoint / relative VR keypoint position & $1.8$ / $0.5$\\
Tracking & Feet / relative feet position & $2.0$ / $0.5$\\
Tracking & Root position / root orientation & $2.0$ / $0.2$--$0.25$\\
Tracking & Body orientation / velocity / angular velocity & $0.5$ / $0.5$ / $0.5$\\
Tracking & Joint position / velocity & $1.0$ / $0.5$\\
Tracking & Contact mask / feet air time & $0.5$ / $1.0$\\
Regularization & Torque / DOF velocity / DOF acceleration & $-10^{-6}$ / $-10^{-4}$ / $-2.5{\times}10^{-8}$\\
Regularization & Action rate / action smoothness & $-0.1$ / $-0.001$\\
Regularization & Foot contact force / stumble / slippage & $-0.01$ / $-2.0$ / $-1.0$\\
Regularization & DOF position / velocity / torque limits & $-10.0$ / $-1.0$ / $-0.5$\\
Regularization & Termination / collision & $-200.0$ / $-30.0$\\
\bottomrule
\end{tabular}}
\end{table*}

\subsection{Motion Sampling Strategy}
\label{subsec:sampling}

We divide each reference into one-second segments and combine offline distribution-aware coverage with online failure-aware prioritization.

\begin{figure}[!t]
  \centering
  \includegraphics[width=\linewidth]{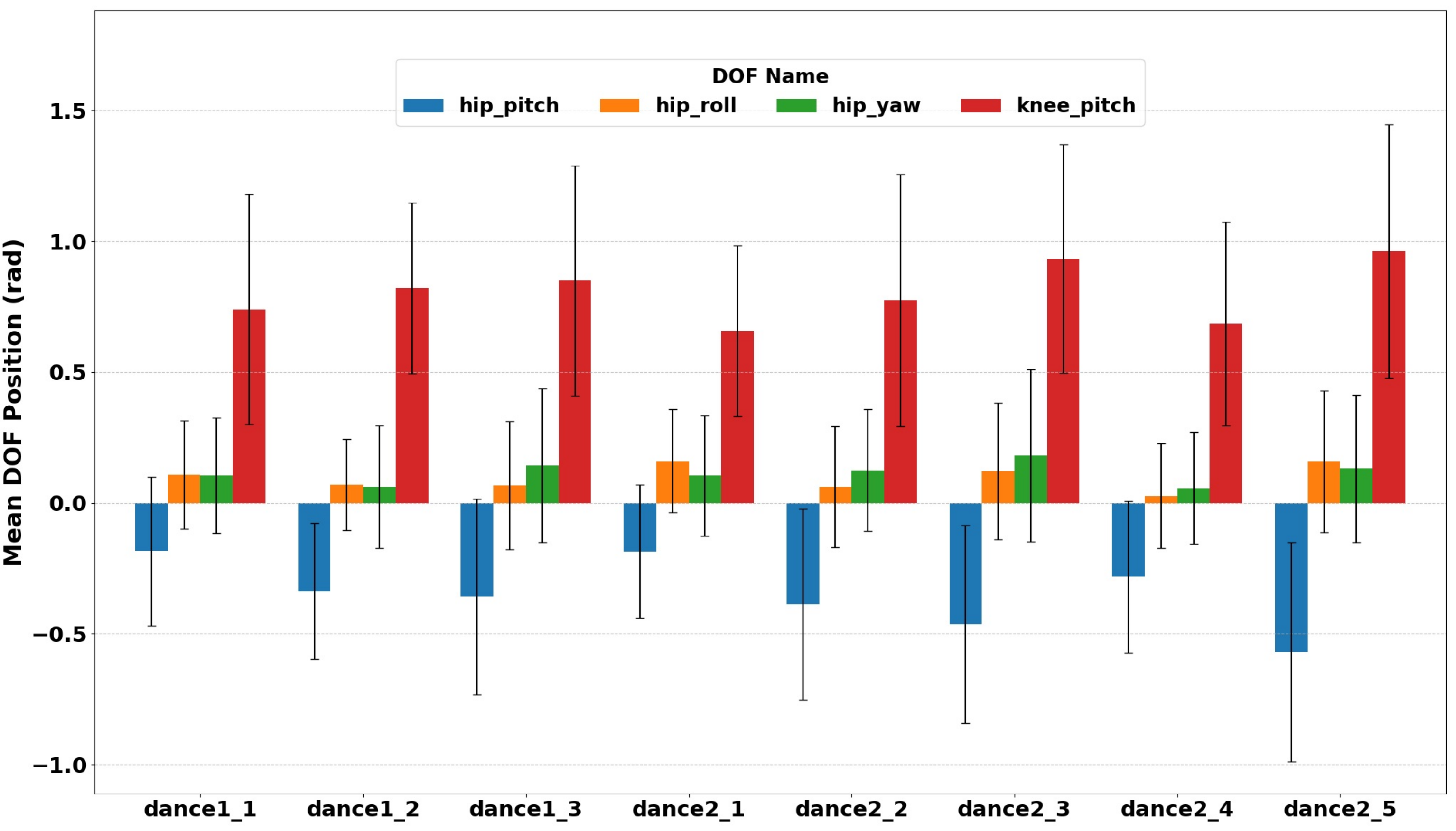}
  \caption{\textbf{Lower-Body Joint Statistics.} Dance segments exhibit substantial variation in key lower-body joint positions.}
  \label{fig:motionbin}
\end{figure}

\begin{figure}[!t]
  \centering
  \includegraphics[width=\linewidth]{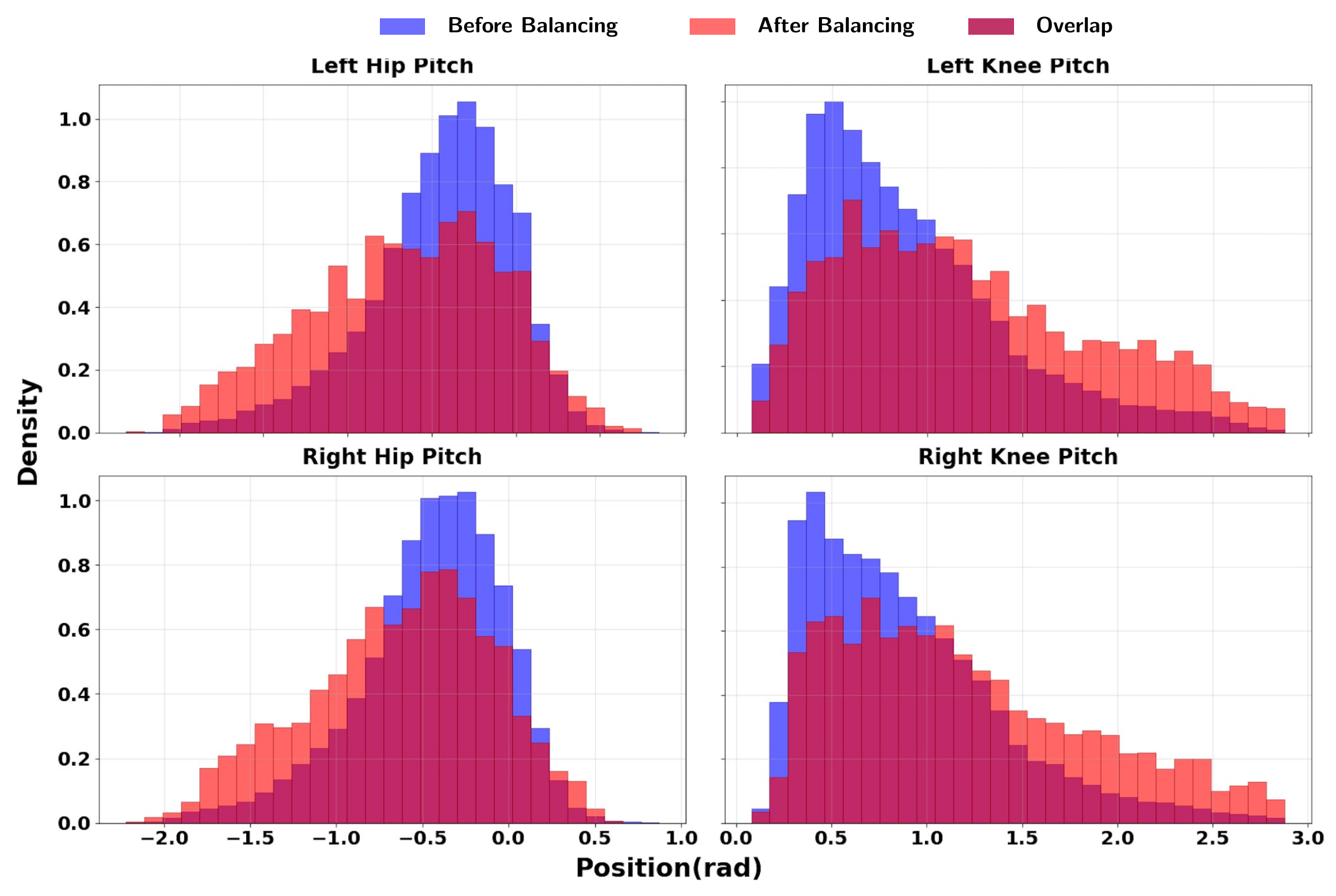}
  \caption{\textbf{Distribution-Aware Balancing.} \textcolor{blue}{Before Balancing} shows the original distribution, \textcolor{red}{After Balancing} shows the balanced distribution, and their overlap produces the third visible color.}
  \label{fig:datadist}
\end{figure}

\textbf{Distribution-Aware Balancing.}
Unlike near-cyclic locomotion~\cite{siekmann2021sim}, long-horizon dance contains heterogeneous configurations across motions and temporal segments. As shown in Fig.~\ref{fig:motionbin}, hip- and knee-pitch joints exhibit particularly large dispersion, while many other lower-body joints remain closer to their neutral values. Uniform time sampling therefore over-represents common configurations, motivating balancing in this compact joint subspace to emphasize rare but informative motion regions.
For each segment $s$, we form a normalized occupancy histogram $\mathbf{h}_s$ over the hip/knee-pitch subspace and stack the histograms as $\mathbf{H}=[\mathbf{h}_1,\ldots,\mathbf{h}_S]$.
The balanced prior is obtained by
\begin{equation}
  \mathbf{w}^{*}=
  \arg\min_{\mathbf{w}\in\Delta^S}
  \|\mathbf{H}\mathbf{w}-\mathbf{U}\|_2^2,
\end{equation}
where $\mathbf{U}$ is a uniform target and $\Delta^S$ is the probability simplex.
We set $p_s^{\mathrm{bal}}=w_s^{*}$, increasing coverage of rare configurations while retaining a uniform exploration floor.
As shown in Fig.~\ref{fig:datadist}, this weighting flattens the sampled marginal distribution of each selected joint, down-weighting common modes and increasing coverage of rare configurations.

\textbf{Failure-Aware Prioritization.}
For segment $s$, we maintain an EMA failure rate
\[
  r_s^{(t+1)}=(1-\alpha)r_s^{(t)}+\alpha f_s^{(t)},
  \qquad f_s^{(t)}\in\{0,1\},
\]
and define
\begin{equation}
  p_s^{\mathrm{fail}}=
  \frac{(r_s+\epsilon)^\beta}
  {\sum_{j=1}^{S}(r_j+\epsilon)^\beta},
  \label{eq:temp}
\end{equation}
where $\beta$ controls prioritization strength.

\textbf{Bounded Mixture.}
The final sampler retains uniform exploration and mixes the balanced and failure priors with ramped weights:
\begin{equation}
\begin{aligned}
  \widetilde p_s(t)={}&
  \lambda_u p_s^{\mathrm{uni}}+
  \lambda_d(t)p_s^{\mathrm{bal}}+
  \lambda_f(t)p_s^{\mathrm{fail}},\\
  p_s(t)={}&
  \mathrm{Norm}\!\left[
  \mathrm{clip}\!\left(\widetilde p_s(t),0,p_{\max}\right)\right].
\end{aligned}
\end{equation}
We use $\lambda_u=0.3$ and ramp $(\lambda_f,\lambda_d)$ from $(0.4,0.2)$ to $(0.55,0.05)$ after a 48K-step warmup over a further 48K steps. The cap prevents collapse onto a few segments. Final conclusions use deterministic fixed-start evaluation, independent of the training sampler.

\textbf{Reference State Initialization.}
We follow RSI~\cite{deepmimic2018}: each episode starts from a sampled reference state with small pose and velocity perturbations.

\subsection{Residual Action Learning}
\label{subsec:residual}

\textbf{Selective Residualization.}
Reference dispersion is used only as a low-cost candidate-selection signal, not as proof of dynamics sensitivity.
For joint $j$, let $\sigma_j=\mathrm{Std}(q^{\mathrm{ref}}_{:,j})$.
We apply a binary mask $\mathbf{m}$:
\begin{equation}
  \mathbf{q}^{\mathrm{tar}}_t=
  \mathbf{q}^{\mathrm{ref}}_{t+1} + 
  \mathbf{m}\odot\mathbf{a}_t .
\end{equation}
The deployed masks select bilateral hip- and knee-pitch joints: G1 uses indices $[0,3,6,9]$, H1 uses $[2,3,7,8]$, and H1-2 uses $[1,3,7,9]$. These joints have large dance-motion excursions and strongly affect support and center-of-mass regulation. Ankle pitch is excluded because ground contact and retargeting mismatch can make corrections around its reference destabilizing. Torso/arm residuals may reduce local joint error but can reduce global robustness. We therefore treat the mask as a stability-oriented heuristic.

\textbf{Feasibility and Smoothness.}
Residuals are bounded per DOF, and commanded positions and torques are projected to platform-specific position, velocity, and torque limits.

\subsection{Curriculum and Robustness}
\label{subsec:curriculum}

Following prior work~\cite{pbhc2025,asap2025}, training progressively tightens tracking tolerances and increases regularization. Domain randomization perturbs physical and actuation parameters before zero-shot deployment.

\begin{table*}[t]
\centering
\small
\setlength{\tabcolsep}{4pt}
\caption{Representative Control, Termination, Randomization, and Resampling Settings.}
\label{tab:method_robustness_settings}
{
\begin{tabular}{lll}
\toprule
Component & Term & Value\\
\midrule
Control & Action scale / clip / torque clipping & $0.25$ / $100$ / enabled\\
Termination & Motion end / motion far / orientation error & enabled / enabled / enabled\\
Domain randomization & Push interval / maximum velocity & $[5,10]$\,s / $0.5$\,m/s\\
Domain randomization & Base COM / link mass factor & $[-0.1,0.1]$\,m / $[0.9,1.2]$\\
Domain randomization & PD-gain / friction factor & $[0.85,1.15]$ / $[0.5,1.25]$\\
Domain randomization & Torque RFI / control delay & $0.1\times[0.5,1.5]$ / $[0,2]$ steps\\
Resampling & Segment / uniform floor & $1$\,s / $0.3$\\
Resampling & Failure/distribution mix & $0.4/0.2\rightarrow0.55/0.05$; 48K warmup/ramp\\
\bottomrule
\end{tabular}}
\end{table*}

\section{EXPERIMENTS}
\label{sec:exp}

\noindent\textit{Scope.}
We train one policy per reference sequence while reusing a common training and deployment recipe.
Unless stated otherwise, results use Unitree G1; H1 and H1-2 are included in the cross-platform study.

\begin{table*}[t]
  \centering
  \caption{Cross-Motion Tracking Errors for the Residual-Action Comparison (mean$\pm$std over 24 evaluation rollouts). Lower is better.}
  \label{tab:residual_comparison}
  \setlength{\tabcolsep}{6pt}
  \renewcommand{\arraystretch}{1.2}
  \begin{tabular}{lccccc}
  \toprule
  Residual Action & $E_{g\text{-}mpbpe}\!\downarrow$ & $E_{mpbpe}\!\downarrow$ & $E_{mpjpe}\!\downarrow$ & $E_{mpjve}\!\downarrow$ & $E_{mpbve}\!\downarrow$ \\
  \midrule
  NONE(baseline)                & 574.68$\pm$152.353 & 109.30$\pm$12.012 & 1967.98$\pm$242.032 & 229.92$\pm$15.467 & 11.73$\pm$0.851 \\
  ALL                 &
  548.76$\pm$132.643 &
  101.09$\pm$14.036 & 1730.13$\pm$236.180 & \textbf{222.63}$\pm$18.717 & 11.33$\pm$1.027 \\
  SELECTIVE (ours)    & \textbf{484.72}$\pm$147.367 & \textbf{89.41}$\pm$14.102 & \textbf{1564.00}$\pm$184.299 & 225.79$\pm$14.014 & \textbf{11.22}$\pm$1.025 \\
  \bottomrule
  \end{tabular}
\end{table*}

\subsection{Implementation Details}
\label{subsec:impl}

\textbf{Data and Platforms.}
We use eight complete LAFAN1 dance references~\cite{lafan,Lv2025LAFAN1Retargeting} retargeted to G1, H1, and H1-2.
All platforms use the same observation/reward design, with morphology-specific robot configurations.

\textbf{Policy and Training.}
The PPO actor and critic use ReLU MLPs with hidden dimensions $[512,256,128]$.
The action dimensions are 23/21/19 for G1/H1-2/H1.
Further reward, control, termination, and randomization settings are given in Sec.~\ref{subsec:curriculum} and the reproducibility appendix~\ref{sec:reproducibility}.

\textbf{Evaluation Protocols.}
We use two complementary evaluation protocols. Cross-motion evaluation aggregates tracking errors, temporal drift, and hardware completion across all eight references to assess performance over motion diversity. Single-motion controlled evaluation isolates method differences on a shared reference and quantifies training-seed uncertainty. These policies are compared at a frozen 10K checkpoint over three independent seeds. Fixed-grid evaluation starts deterministic rollouts every 1\,s along the reference, while start-zero evaluation measures end-to-end survival. Each seed is reduced to one statistic before reporting the seed-level mean and 95\% Student-$t$ confidence interval; fixed starts are not treated as independent training runs.

\textbf{Metrics.}
Following~\cite{pbhc2025}, we report global and root-relative mean per-body position errors
$E_{g\text{-}mpbpe}$ and $E_{mpbpe}$ (mm), mean per-joint position error
$E_{mpjpe}$ (mrad), and their velocity counterparts $E_{mpbve}$ and $E_{mpjve}$.
The single-motion studies additionally report fixed-grid completion and mean survival, plus start-zero survival where relevant.

\subsection{Aggregate Cross-Motion Evaluation}
\label{subsec:ablation}

\begin{figure}[t]
  \centering
  \includegraphics[width=\linewidth]{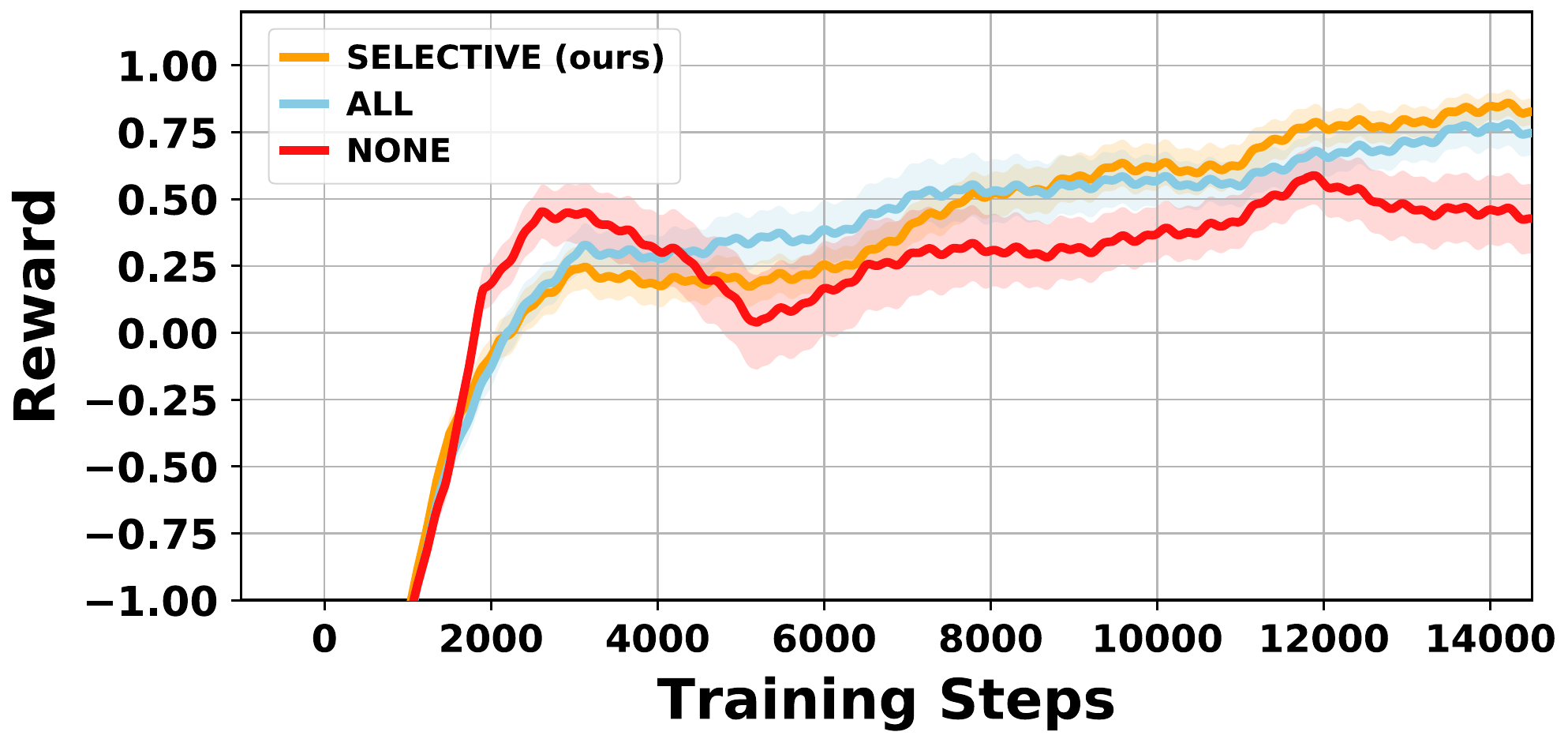}
  \caption{\textbf{Learning Curves under Residualization Strategies.} Representative single-reference diagnostic showing the mean over three training seeds with 95\% confidence-interval shading.}
  \label{fig:learning_curve}
\end{figure}

\textbf{Residual Actions.}
The cross-motion comparison uses NONE (absolute actions), ALL (all-DOF residuals), and SELECTIVE (hip/knee-pitch residuals).
Across the aggregate in Table~\ref{tab:residual_comparison}, SELECTIVE reduces
$E_{g\text{-}mpbpe}$, $E_{mpbpe}$, and $E_{mpjpe}$ by 15.7\%, 18.2\%, and 20.5\% relative to NONE.
It also improves most metrics over ALL.
The consistent position-error reductions across the eight-motion aggregate support that selective residualization improves tracking fidelity over diverse references, while the $E_{mpjve}$ exception avoids claiming superiority on every smoothness metric.
Fig.~\ref{fig:learning_curve} summarizes the corresponding residual-learning dynamics.
NONE rises quickly but then oscillates and settles at a lower return; ALL improves earlier, whereas SELECTIVE overtakes it during later training and reaches the highest stable plateau. This supports selective residual authority as a mechanism for stable late-stage refinement rather than merely faster initial learning.

\begin{figure}[t]
  \centering
  \includegraphics[width=\linewidth]{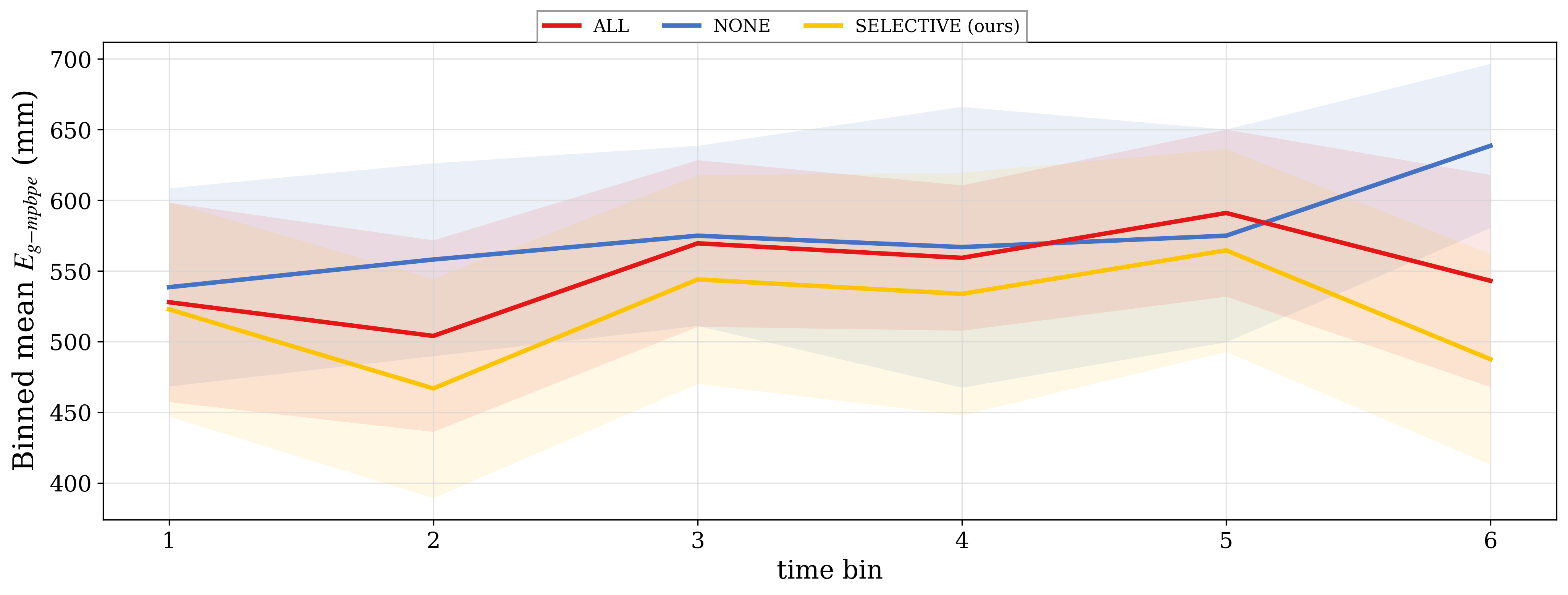}
  \caption{\textbf{Cross-Motion Time-binned Tracking Error.} Mean $E_{g\text{-}mpbpe}$ across the six relative temporal bins; shading denotes the 95\% bootstrap confidence interval over the evaluation aggregate.}
  \label{fig:error_drift}
\end{figure}

\textbf{Cross-Motion Error Accumulation.}
For Fig.~\ref{fig:error_drift}, each complete reference rollout is divided into six temporal bins before aggregation across motions, seeds, and repeated deterministic rollouts. SELECTIVE remains below NONE across the motion progression and avoids its pronounced final-bin degradation. Together with Table~\ref{tab:residual_comparison}, this supports reduced long-horizon error accumulation rather than only improved short-term motion tracking.

\begin{figure}[t]
  \centering
  \includegraphics[width=\linewidth]{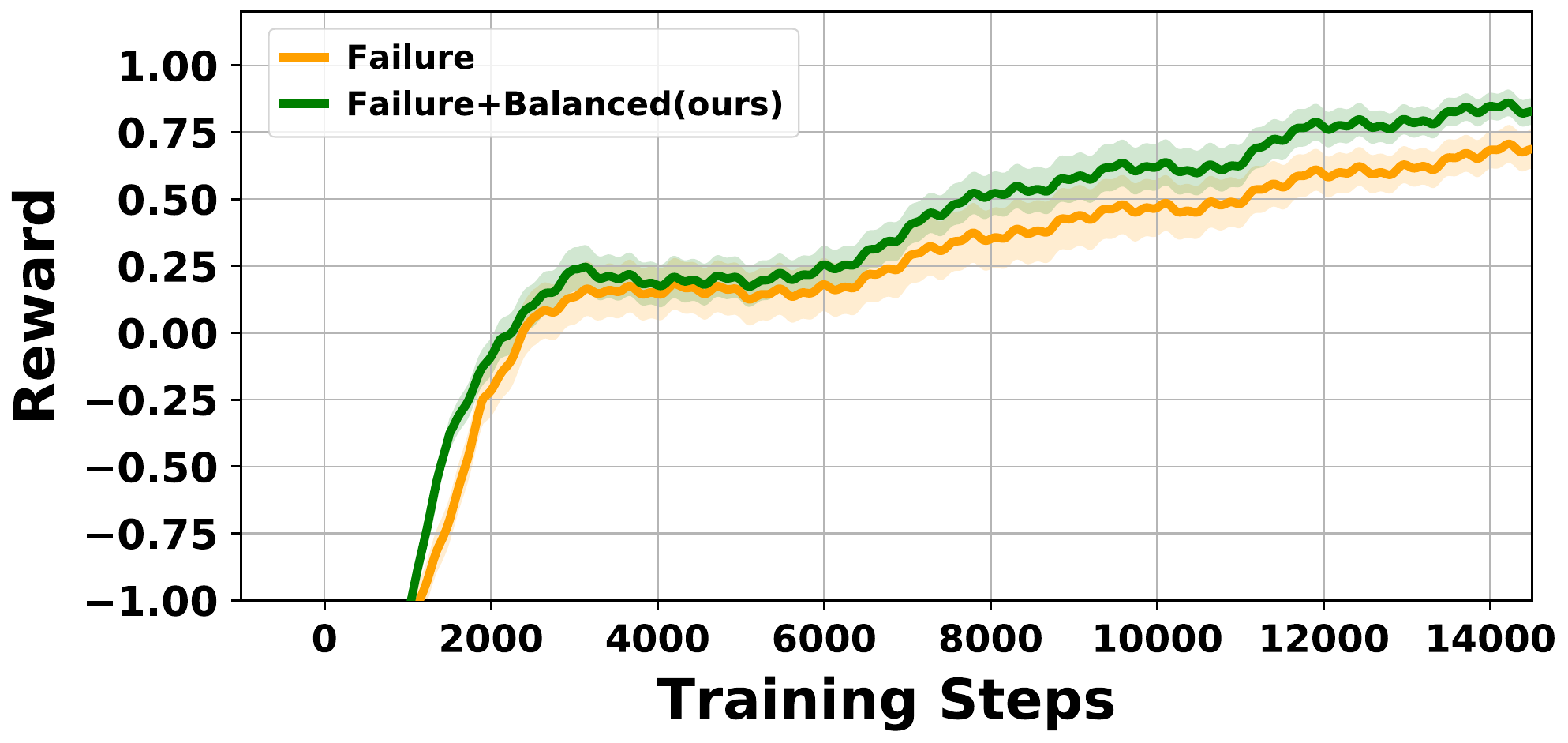}
  \caption{\textbf{Learning Curves for Sampling Variants.} Representative single-reference diagnostic showing the mean over three training seeds with 95\% confidence-interval shading.}
  \label{fig:sampling_curve}
\end{figure}

\textbf{Sampling Diagnostic.}
Fig.~\ref{fig:sampling_curve} compares failure-aware replay with its distribution-balanced extension; Table~\ref{tab:r5_resampling_ablation} reports the four-condition fixed-budget results.
Failure + Balanced improves earlier and remains above Failure through the final training stage, indicating that distribution-aware coverage complements failure replay and improves sample efficiency without replacing hard-example prioritization.

\subsection{Cross-Motion Hardware Performance}
\label{subsec:performance}

We evaluate three G1 hardware trials for each of the eight references.
All references complete in at least two trials, yielding 21/24 successful trials (87.5\%).
Policies are trained in simulation, validated in MuJoCo, exported to TorchScript, and executed on the onboard Orin NX at 50\,Hz following~\cite{margolis2023walk}, without test-time filtering or action rescaling.

\begin{table}[t]
  \centering
  \begin{threeparttable}
    \caption{G1 Hardware Evaluation across Sequences.}
    \label{tab:dance_catalog}
    \setlength{\tabcolsep}{6pt}
    \renewcommand{\arraystretch}{1.08}
    \begin{tabular}{lcc}
      \toprule
      \textbf{Name (LAFAN1)} & \textbf{Sim (MuJoCo)} & \textbf{Real (G1)} \\
      \midrule
      \text{dance1\_subject1} & success$^{1}$ & success (2/3) \\
      \text{dance1\_subject2} & success & success (3/3) \\
      \text{dance1\_subject3} & success & success (3/3) \\
      \text{dance2\_subject1} & success & success (2/3) \\
      \text{dance2\_subject2} & success & success (3/3) \\
      \text{dance2\_subject3} & success & success (3/3) \\
      \text{dance2\_subject4} & success & success (2/3) \\
      \text{dance2\_subject5} & success & success (3/3) \\
      \bottomrule
    \end{tabular}
    \begin{tablenotes}[flushleft]
      \item $1$ Success means completing the full reference without a fall, safety stop, or disallowed body contact.
    \end{tablenotes}
  \end{threeparttable}
\end{table}

\begin{figure}[t]
  \centering
  \includegraphics[width=\linewidth]{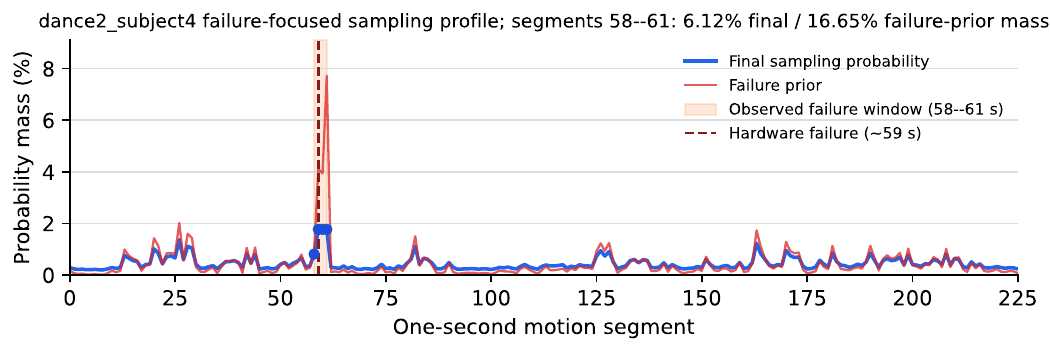}
  \caption{\textbf{Failure-Aware Sampling Profile on \texttt{dance2\_subject4}.} The hardware failure near 59\,s lies in the highlighted 58--61\,s difficult window.}
  \label{fig:r4_failure_sampling_profile}
\end{figure}

\textbf{Failure Analysis.}
The three unsuccessful trials occurred once each on \texttt{dance1\_subject1}, \texttt{dance2\_subject1}, and \texttt{dance2\_subject4}. The clearest case failed at approximately 59\,s of \texttt{dance2\_subject4}, during a low, contact-rich floor-to-stand transition from which the robot did not recover. In a representative combined-sampling run, segment 58 has the highest EMA failure rate (94.5\%), while segments 59--61 receive the capped maximum sampling probability (1.77\% each); Fig.~\ref{fig:r4_failure_sampling_profile} shows this alignment.

\subsection{Single-Motion Controlled Studies}
\label{subsec:controlled_studies}

\textbf{Common-Protocol Baselines.}
Table~\ref{tab:r3_baselines} compares our ASAP and KungfuBot-style reimplementations with RobotDancing on the 131.5\,s \texttt{dance1\_subject2} reference. All use the same simulator, retargeted reference, termination protocol, and three seeds. RobotDancing completes the start-zero sequence across seeds and achieves substantially higher fixed-grid completion and survival. We avoid comparing tracking errors after unequal termination horizons. The ASAP and KungfuBot-style entries are our reimplementations rather than values reported by the cited papers.

\begin{table}[!t]
\centering
\footnotesize
\setlength{\tabcolsep}{3pt}
\caption{Baselines Comparison on \texttt{dance1\_subject2}. Values use 3 seeds and 132 fixed starts per seed.}
\label{tab:r3_baselines}
{
\resizebox{\columnwidth}{!}{%
\begin{tabular}{lccc}
\toprule
Method & Fixed comp. (\%) $\uparrow$ & Fixed surv. (s) $\uparrow$ & Start-zero (s) $\uparrow$\\
\midrule
ASAP-style & 3.3$\pm$1.1 & 2.5$\pm$0.5 & 5.7$\pm$0.2\\
KungfuBot-style & 4.3$\pm$2.9 & 3.0$\pm$0.4 & 13.5$\pm$6.4\\
RobotDancing & \textbf{97.7$\pm$1.9} & \textbf{64.6$\pm$0.6} & \textbf{131.5$\pm$0.0}\\
\bottomrule
\end{tabular}}}
\end{table}

\textbf{Residual-Mask Ablation.}
\texttt{Dance2\_subject4} study in Table~\ref{tab:r5_masks} separates robustness from local tracking accuracy. SELECTIVE gives the highest completion and survival while keeping MPBPE/MPJPE below the absolute-action baseline. Adding ankle or broader residual authority may lower local MPJPE, but can reduce robustness or increases global body error; the mask is therefore a stability-oriented heuristic rather than an accuracy-only optimum.

\begin{table}[!t]
\centering
\footnotesize
\setlength{\tabcolsep}{2.5pt}
\caption{Residual-Mask Ablation on \texttt{dance2\_subject4}. Values use 3 seeds and 226 fixed starts per seed.}
\label{tab:r5_masks}
{
\resizebox{\columnwidth}{!}{%
\begin{tabular}{lcccc}
\toprule
Action / mask & Comp. (\%) $\uparrow$ & Surv. (s) $\uparrow$ & MPBPE $\downarrow$ & MPJPE $\downarrow$\\
\midrule
Absolute & 23.6$\pm$3.9 & 37.2$\pm$15.5 & 445.4$\pm$55.3 & 174.0$\pm$12.4\\
Selective (Ours) & \textbf{38.6$\pm$22.6} & \textbf{45.0$\pm$15.7} & 424.2$\pm$37.4 & 163.9$\pm$3.4\\
Hip/Knee/Ankle & 17.8$\pm$3.9 & 34.1$\pm$13.9 & 489.1$\pm$107.7 & 159.7$\pm$19.1\\
All Legs & 20.8$\pm$28.3 & 31.5$\pm$25.6 & 418.0$\pm$45.5 & 158.1$\pm$10.8\\
Legs+Torso & 20.8$\pm$7.2 & 33.4$\pm$8.5 & \textbf{415.6$\pm$64.0} & 152.5$\pm$14.5\\
All-DOF & 24.3$\pm$74.8 & 37.0$\pm$49.5 & 682.0$\pm$190.7 & \textbf{124.6$\pm$42.4}\\
\bottomrule
\end{tabular}}}
\end{table}

\textbf{Motion-Resampling Ablation.}
\texttt{dance1\_subject2} saturates for Failure-only and Combined, so Table~\ref{tab:r5_resampling_ablation} uses the longer \texttt{dance2\_subject4} reference. Distribution-only improves early coverage, Failure-only becomes stronger at later checkpoints, and Combined integrates both effects and reaches the strongest threshold-success and survival results.

\begin{table}[!t]
\centering
\scriptsize
\setlength{\tabcolsep}{3pt}
\caption{Resampling Ablation on \texttt{dance2\_subject4}. Values are seed-level mean$\pm$95\% CI over 3 seeds and 226 fixed starts.}
\label{tab:r5_resampling_ablation}
\label{tab:r5_resampling_checkpoint}
{
\textbf{(a) Threshold Success.}\\[2pt]
\begin{tabular}{lccc}
\toprule
Sampler & 10\,s & 20\,s & 30\,s\\
\midrule
Uniform RSI & 62.2$\pm$10.2\% & 29.9$\pm$12.3\% & 10.0$\pm$9.2\%\\
Distribution-only & 65.5$\pm$3.8\% & 37.3$\pm$4.2\% & 18.6$\pm$4.0\%\\
Failure-only & 69.8$\pm$13.4\% & 40.4$\pm$21.6\% & 22.0$\pm$24.5\%\\
Combined (Ours) & \textbf{72.4$\pm$3.4\%} & \textbf{45.7$\pm$4.4\%} & \textbf{26.1$\pm$6.7\%}\\
\bottomrule
\end{tabular}

\vspace{3pt}
\textbf{(b) Mean Survival Progression (s).}\\[2pt]
\begin{tabular}{lccc}
\toprule
Sampler & 4K & 6K & 10K\\
\midrule
Uniform RSI & 14.1$\pm$1.6 & 13.4$\pm$2.9 & 15.5$\pm$4.3\\
Distribution-only & 14.4$\pm$0.5 & 16.3$\pm$2.0 & 18.5$\pm$1.8\\
Failure-only & 13.9$\pm$0.8 & 16.0$\pm$5.0 & 19.9$\pm$9.2\\
Combined (Ours) & 14.0$\pm$1.7 & 16.2$\pm$1.8 & \textbf{21.6$\pm$2.7}\\
\bottomrule
\end{tabular}}
\end{table}

\subsection{Cross-Platform Validation}
\label{subsec:cross}

Table~\ref{tab:r4_platform_final} reports a controlled sim-to-sim pilot on the same \texttt{dance1\_subject2} reference. G1 and H1 achieve high fixed-grid completion and complete the start-zero sequence across seeds. H1-2 remains weaker under strict motion-far termination: its start-zero rollouts stop around 34\,s when the tracking threshold is reached, rather than because of a fall. We therefore interpret the result as strong portability to H1 and a remaining H1-2 tracking-accuracy gap. H1/H1-2 hardware clips, including visible jitter, are qualitative feasibility evidence only.

\begin{table}[!t]
\centering
\scriptsize
\setlength{\tabcolsep}{2.5pt}
\caption{Cross-Platform Simulation on \texttt{dance1\_subject2}. Values use 3 seeds and 132 fixed starts per seed.}
\label{tab:r4_platform_final}
{
\textbf{(a) Fixed-grid Evaluation.}\\[2pt]
\resizebox{\columnwidth}{!}{%
\begin{tabular}{lcccc}
\toprule
Robot & Completion (\%) $\uparrow$ & Survival (s) $\uparrow$ & MPBPE $\downarrow$ & MPJPE $\downarrow$\\
\midrule
G1 & 97.7$\pm$1.9 & 64.6$\pm$0.6 & 1641$\pm$420 & 118.1$\pm$19.9\\
H1 & 99.7$\pm$1.1 & 65.7$\pm$1.1 & 3492$\pm$634 & 118.7$\pm$2.5\\
H1-2 & 64.4$\pm$29.9 & 35.3$\pm$16.4 & 1327$\pm$580 & 165.9$\pm$7.3\\
\bottomrule
\end{tabular}}

\vspace{3pt}
\textbf{(b) Start-Zero Survival (s).}\\[2pt]
\begin{tabular}{lccc}
\toprule
Robot & G1 & H1 & H1-2\\
\midrule
Survival & 131.5$\pm$0.0 & 131.5$\pm$0.0 & 34.1$\pm$1.0\\
\bottomrule
\end{tabular}}
\end{table}

\section{CONCLUSIONS}
We presented RobotDancing, a single-stage residual-action RL recipe for humanoid motion tracking. By predicting reference-conditioned residual joint targets and combining distribution-aware balancing with failure-aware prioritization, the controller reduces error accumulation and improves training efficiency. Across eight dance references, the results demonstrate long-horizon tracking on G1, zero-shot G1 hardware deployment, and portability in controlled G1/H1/H1-2 simulation studies.
\noindent\textbf{Limitations.}
(i) RobotDancing trains one policy per reference sequence; universal tracking must address multimodal references, cross-motion sampling interference, and motion identity and phase inference. (ii) The morphology-dependent residual mask remains heuristic. (iii) H1/H1-2 hardware evidence is qualitative. (iv) Hardware failure diagnosis is limited by the absence of synchronized torque, contact, slip, and base-attitude telemetry. Future work will study motion-conditioned residual gates, per-motion sampling statistics, and perception-aware tracking in unstructured environments.

\FloatBarrier
\begingroup
\raggedbottom

\section*{APPENDIX: REPRODUCIBILITY DETAILS}
\label{sec:reproducibility}

This appendix complements the reward scales in Table~\ref{tab:method_reward_scales} and the common robustness settings in Table~\ref{tab:method_robustness_settings}. Unless stated otherwise, the settings below are shared across motions. Platform-specific PD gains, joint/velocity/torque limits, and morphology-specific termination thresholds are provided in the released robot configuration files.

\begin{table}[H]
\centering
\footnotesize
\setlength{\tabcolsep}{3pt}
\renewcommand{\arraystretch}{0.94}
\caption{Reward Tolerances and Curriculum.}
\label{tab:app_reward}
{
\begin{tabularx}{\columnwidth}{@{}p{0.44\columnwidth}X@{}}
\toprule
Term & Value\\
\midrule
Upper-/lower-body position $\sigma$ & $0.02/0.02$\\
VR keypoint / feet position $\sigma$ & $0.02/0.015$\\
Root position / orientation $\sigma$ & $0.25/0.25$\\
Body orientation $\sigma$ & $0.10$\\
Body linear / angular velocity $\sigma$ & $1.0/15.0$\\
Joint position / velocity $\sigma$ & $0.30/30.0$\\
Adaptive tracking $\sigma$ & enabled; origin-based; $\alpha=10^{-3}$\\
Penalty scale (initial/min/max) & $0.80/0.80/1.00$\\
Penalty update degree; down/up thresholds & $10^{-5}$; $40/42$\\
Position/velocity/torque limits & $0.90/0.90/0.85$ of platform limits\\
Maximum foot-contact force & $400$\,N\\
\bottomrule
\end{tabularx}}
\end{table}

\begin{table}[H]
\centering
\footnotesize
\setlength{\tabcolsep}{3pt}
\renewcommand{\arraystretch}{0.94}
\caption{PPO, Network, and Training Parameters.}
\label{tab:app_training}
{
\begin{tabularx}{\columnwidth}{@{}p{0.42\columnwidth}X@{}}
\toprule
Term & Value\\
\midrule
Actor / critic MLP & $[512,256,128]$ / $[512,256,128]$; ReLU\\
Rollout length / learning epochs & $24$ steps/env / $5$\\
Mini-batches / PPO clip & $4/0.2$\\
Discount $\gamma$ / GAE $\lambda$ & $0.99/0.95$\\
Value-loss / entropy coefficient & $1.0/0.01$\\
Actor / critic learning rate & $10^{-3}/10^{-3}$\\
Gradient-norm limit & $1.0$\\
Lr schedule / target KL & adaptive / $0.01$\\
Initial action-noise variance & $0.8$\\
Checkpoint interval & $500$ iterations\\
Main G1 training & one RTX 4090; $8192$ envs; seeds $1,2,3$\\
Training budget & final policies: 20K iterations; controlled studies: 10K\\
\bottomrule
\end{tabularx}}
\end{table}

\begin{table}[H]
\centering
\footnotesize
\setlength{\tabcolsep}{3pt}
\renewcommand{\arraystretch}{0.94}
\caption{Control, Initialization, and Termination Parameters.}
\label{tab:app_control}
{
\begin{tabularx}{\columnwidth}{@{}p{0.42\columnwidth}X@{}}
\toprule
Term & Value\\
\midrule
Simulation / policy-control rate & $500/50$\,Hz\\
Action scale / clip & $0.25/100$\\
Command projection & joint-position and torque clipping enabled\\
Action dimension(G1/H1/H1-2) & $23/19/21$\\
Residual indices: G1 & $[0,3,6,9]$\\
Residual indices: H1 / H1-2 & $[2,3,7,8]$ / $[1,3,7,9]$\\
Residual semantics & bilateral hip- and knee-pitch joints\\
RSI joint position/velocity noise & $0.10/0.15$\\
RSI root position/rotation noise & $0.05$\,m / $10^\circ$\\
RSI root lin./ang. vel. noise & $0.01/0.01$\\
Termination conditions & motion end, motion far, orientation error\\
Motion-far threshold & $0.50/0.30/0.50$\,m (init/min/max)\\
Motion-far update degree; down/up thresholds & $10^{-5}$; $40/42$\\
Orientation-error threshold & $0.75$\,rad\\
\bottomrule
\end{tabularx}}
\end{table}

\begin{table}[H]
\centering
\footnotesize
\setlength{\tabcolsep}{3pt}
\renewcommand{\arraystretch}{0.94}
\caption{Domain Randomization and Motion Resampling.}
\label{tab:app_robustness}
{
\begin{tabularx}{\columnwidth}{@{}p{0.42\columnwidth}X@{}}
\toprule
Term & Range / value\\
\midrule
Push interval / max velocity & $[5,10]$\,s / $0.5$\,m/s\\
Base COM offset & $[-0.1,0.1]$\,m per axis\\
Link-mass / friction factor & $[0.9,1.2]/[0.5,1.25]$\\
PD factor & $[0.85,1.15]/[0.85,1.15]$\\
Torque RFI & nominal $0.1$; factor $[0.5,1.5]$\\
Control delay & $[0,2]$ control steps\\
Segment duration & $1$\,s\\
Resampling dist. strength & $0.5$\\
Resampling uniform floor & $0.3$\\
Failure EMA $\alpha$ / $\epsilon$ / $\beta$ & $0.05/10^{-3}/2.0$\\
Failure-statistics update interval & $100$ steps\\
Init fail/dist mixture & $0.40/0.20$\\
Final fail/dist mixture & $0.55/0.05$\\
Mixture handoff start / ramp & $48$K / $48$K steps\\
Per-segment probability cap & $4\times$ uniform probability\\
\bottomrule
\end{tabularx}}
\end{table}

\FloatBarrier
\endgroup

{\small
\bibliographystyle{IEEEtran}
\bibliography{reference}
}

\end{document}